# IEEE Copyright Notice

# Application of Federated Machine Learning in Manufacturing

1st Vinit Hegiste
*Chair of Machine Tools and Control Systems*
*TU Kaiserslautern*
Kaiserslautern, Germany
vinit.hegiste@mv.uni-kl.de

2nd Tatjana Legler
*Chair of Machine Tools and Control Systems*
*TU Kaiserslautern*
Kaiserslautern, Germany
tatjana.legler@mv.uni-kl.de

3rd Martin Ruskowski
Innovative Factory Systems (IFS)
*German Research Center for Artificial Intelligence (DFKI)*
Kaiserslautern, Germany
martin.ruskowski@dfki.de

*Abstract*—A vast amount of data is created every minute, both in the private sector and industry. Whereas it is often easy to get hold of data in the private entertainment sector, in the industrial production environment it is much more difficult due to laws, preservation of intellectual property, and other factors. However, most machine learning methods require a data source that is sufficient in terms of quantity and quality. A suitable way to bring both requirements together is federated learning where learning progress is aggregated, but everyone remains the owner of their data.

Federate learning was first proposed by Google researchers in 2016 and is used for example in the improvement of Google's keyboard Gboard. In contrast to billions of android users, comparable machinery is only used by few companies. This paper examines which other constraints prevail in production and which federated learning approaches can be considered as a result.

## I. INTRODUCTION

Machine Learning has become part of many sectors, including our daily life. Ranging from Netflix recommendations [1] to smart healthcare [2], fraud detection [3], earthquake prediction [4], recycling and waste management [5], and the development of self-driving cars [6]. The possible applications of ML in production and manufacturing are similarly diverse [7], [8].

Besides the production of defects, the prevention and implementation of quality measures are the cost for preventing and removing defects, and costs for quality measures to be considered in total quality management [9]. ML-based approaches can be used to monitor processes or predict the product quality [10], [11]. Through the recording and analysis of the operational data of machinery, predictive maintenance helps minimise unplanned downtime and optimize tool life [10]. New insights can be gained through the use of additional sensors in machines. For example, [12] used microphones to detect chatter on a CNC lathe and thus helps to ensure optimal process parameters. Product quality control with CNN has higher adaptability to changing conditions and therefore reduces rejection rates [13], [14].

The widespread use and high accuracy of ML models can be attributed to the high availability of data, for example, from mobile or IoT devices [15]. ML as-a-service providers, such as IBM Cloud, Amazon Web Services, Google Cloud, and Microsoft Azure [16], compute the models on their high-performance servers. In this centralised architecture all data is sent to the cloud and the client sends a request to the server. Although the achievable accuracy is the highest in this case, data privacy concerns are increasing and stricter regulations such as General Data Protection Regulation (GDPR) came into force [17]. This led to a decentralised, on-device machine learning approach [18].

## II. RELATED WORK

### A. Federated learning

Google introduced federated learning in 2016 [19]. They developed a new concept called Federated optimisation, with the goal of creating a high-quality centralised global model based on models trained on data from unevenly distributed data across an extremely large number of nodes (edge devices, client's machines, smart phones, etc). It is a novel technique that enables multiple users to train a global model based on local weights obtained from the local data of the users participating in the training round. By iterating this process, the global model achieves a high accuracy based on the local data from various users/devices used to train this model. The main advantage of this technique is that raw sensitive data or image never ever leaves the user's device, and only the model weights are sent to the global server. This helps to reduce privacy concerns when training a model on sensitive data such as health care, personal chats, sensor data from manufacturing companies, and so on. Figure 1 depicts a generic diagram of the federated learning architecture, in which the machine learning model is trained on the user's device, which could be a laptop, edge device, computer, or even a smartphone.

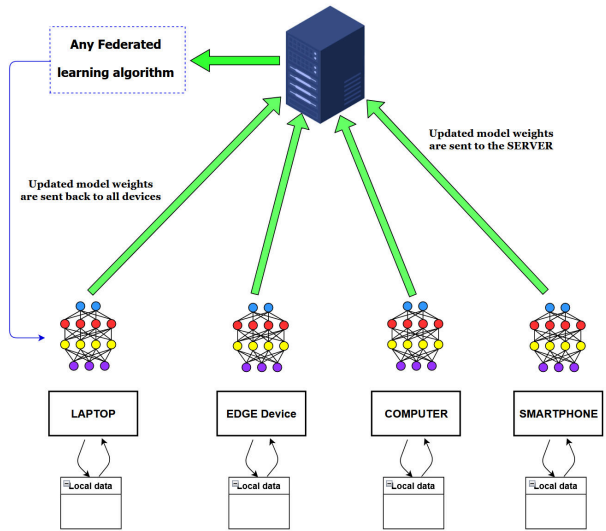

Fig. 1. Federated learning with multiple users/devices

After training, the model weights are sent back to the server, which uses a federated learning algorithm/framework such as FedAvg [19], FedProx [20], FedMA [21], FedVision [22], etc. to generate a global federated model, which is then sent back to all users for the next update. This single iteration of sending a global model to users and receiving weights from all users before applying a federated learning algorithm to obtain the global model is referred to as a single communication round (CR).

Google uses this federated learning technique in their application called Gboard (Google keyboard), where the data of users who use this application is stored locally on their devices [23]. Google sends a global model to these devices, which train it with local data and send the updated weights and gradients back to the Google server. On the server side, an algorithm called FedAvg [19] is used, which computes the average of these weights (from all clients) and updates the model weights with these averaged weights. This revised model is returned to the clients for training on newly generated data. The question of methodology arises, because in this case, the local devices require computing power and memory, as well as frequent communication between the devices and the server, which will require high bandwidth in order to exchange machine learning model parameters. Google finds a solution to this problem by training the model on mobile devices while they are charging, connected to Wi-Fi, and not in use. Only under the aforementioned conditions are model parameters sent and a new update model is downloaded for training on local data [23].

### B. Basic architecture and Algorithms

McMahan et al. mentioned two algorithms in [19], namely FedSGD and FedAvg. In FedSGD, the clients train the entire local data set over one epoch and then send the updated local weight gradients to the central server. This server then averages out all of the weight gradients from all of the clients, applies the weight update, and sends the updated weights to the clients. The important point to note here is that in this entire process, all clients locally take one step of gradient descent, as shown in figure 2.

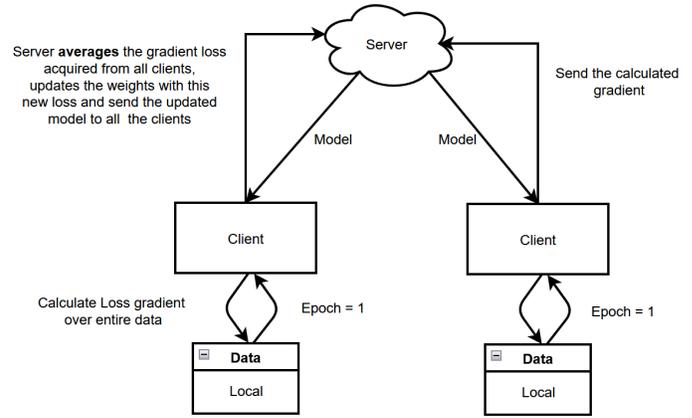

Fig. 2. Federated SGD [19]

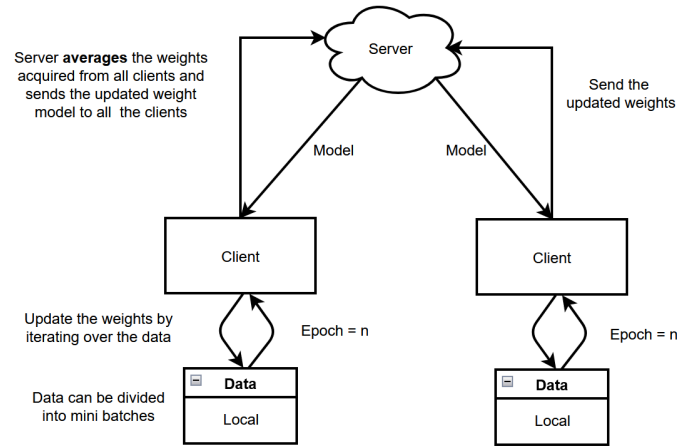

Fig. 3. Federated Averaging [19]

FedSGD algorithm provides as a base to FedAvg algorithm where more computation is performed on the local side such that each client iterates multiple times before the averaging step i.e. increasing the epochs and introducing mini batches as shown in figure 3. This paper concluded that with FedAvg algorithm the number of communication rounds between a server and all clients can be reduced by properly hyper-tuning the number of local epochs and the batch size for all the clients. The main issue with federated learning is the communication rounds, as these model weights must be transferred from all clients to the server and back again, and in real life, problems such as internet outage, power outage, and so on can impede the execution of a single communication round. As a result, having hundreds of such communication rounds increases the algorithm's failure in an uncontrolled environment. Thus, increasing the epochs on the local client side while decreasing the number of communication rounds is an effective solution.

FedAvg serves as the foundation for frameworks/algorithms such as FedProx [20], FedMA [21], FedVision [22], etc. Additionally, this federated learning can be subdivided into a number of categories [24], out which we use Horizontal federated learning [25]. Horizontal federated learning (also called as sample-based federated learning) is used in scenario when data sets share the same features but are different in samples [25]. Our use case has the similar scenario (refer section III-A) where the clients' data set share similar features but are different in samples.

## III. Federated learning in Manufacturing for Quality inspection

The introduction of federated learning is advantageous for businesses, research institutions, and even individuals attempting to collaborate on applications such as quality inspection, anomaly detection, object detection, etc. This technique can be useful for inter-company applications in which two or more companies train a model collaboratively, as well as intra-company applications in which a company with multiple branches or production locations wishes to train a deep learning model for any of the aforementioned applications. Using horizontal federated learning [25], we constructed a USB quality inspection model for our use case. This use case aims to construct a global USB classification model capable of classifying all mechanical forms of USB port errors/faults present in the data set. This use case includes three clients with their own local data, as depicted in the figure 4, which are trained using a VGG-19 model [26]. We have experimented with this entire data set trained using VGG-19 model and a global federated model trained using 3 clients (the total data used is same but each USB sticks' data set is represented as an individual client), where the starting weights used are pretrained ImageNet weights [27]. In regards to federated machine learning, we also conducted experiments to determine the difference in the total number of Communication Rounds required by the federated learning algorithm when using the fine-tuned VGG model and a custom fully-connected layered VGG model (Transfer learning). Our application's workflow adheres to FedAvg [19] (figure 3) because it is the foundation for federation learning and easy to implement.

### A. Data set

Our data set consists of three different types of USB sticks. Figure 4 depicts three different USB stick with four classes each. Each USB stick's data functions as local data for each client, hence in our federated learning scenario we have three clients. Huawei Openlab Munich's [28] USB stick acted as client1's data, SmartFactory-KL's [29] USB acted as client2's data and Chair of Machine Tools and Control Systems's (at Technische Universität Kaiserslautern) [30] USB acted as client3's data. Client1 possesses a hexagonal USB drive with a rotating body that protrudes the USB plug from the body. Client2 has a rectangular, blue USB stick that resembles a LEGO brick, while Client3 has a similar USB stick that is red in color. The distinguishing characteristic of this data set is that the error on each client is unique. Client1 has little stickers on its USB port that represent a scratch error, Client2 has a damaged USB port as its fault, and Client3 has a rusty USB port as its error. The "HIDDEN" class represents a picture in which the USB port is not clearly visible to the camera, making it impossible to determine whether the USB port of the USB stick is damaged/inoperative. And the "EMPTY" class is comprised of blank images and also images of non-USB objects such as wires, screws, etc.

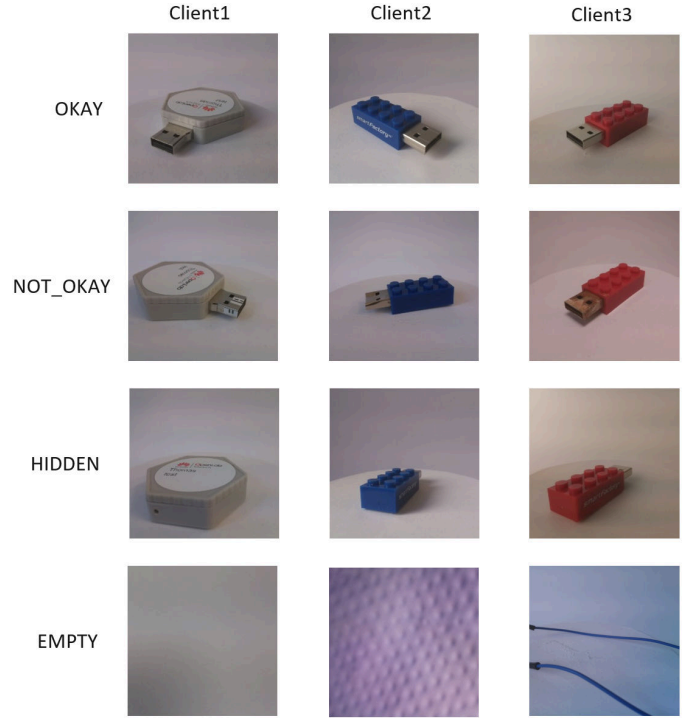

Fig. 4. Different USB sticks and classes belonging to individual clients

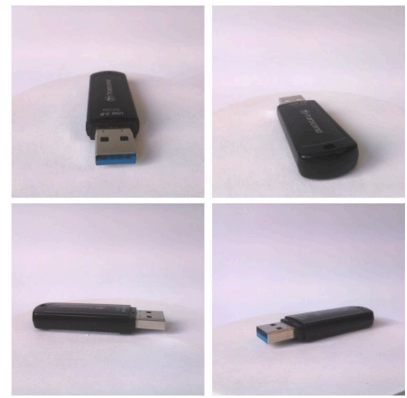

Fig. 5. External data set of USB with 3.0 port for testing purpose

All three Clients share the "OKAY" class, where the camera can clearly observe an error-free USB port. The data set consists of images taken from various angles and different lighting conditions. The total number of images per client is kept unequal (figure 6) to demonstrate real life scenarios in which custom data sets are not always independent and identically distributed i.e. non-IID data set. For testing purpose we also used a USB 3.0 (figure 5) which had no errors to ensure if the final model created (whether federated or normal) is not over-fitted. This USB3.0 data was manually annotated and had a total of 3071 out of which 255 are "HIDDEN" and rest are "OKAY" images.

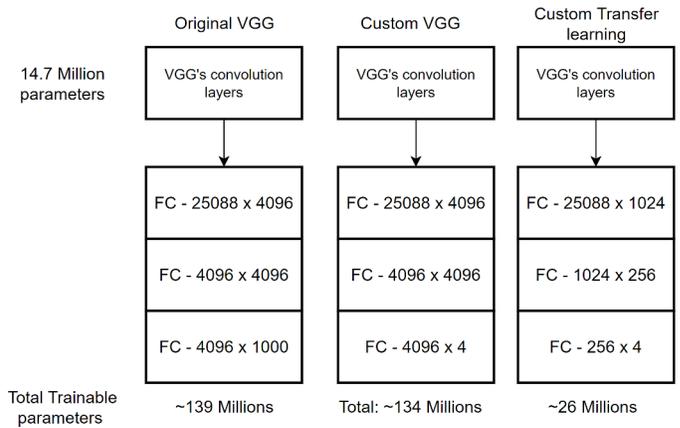

Fig. 7. VGG 19's original fully connected (FC) layers vs Fine-tuned VGG's custom FC layers vs Transfer learning VGG's custom FC layers

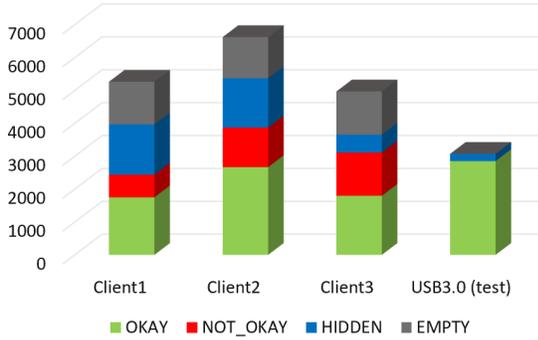

Fig. 6. Custom data distribution

### B. Architecture

VGG-11, VGG-16, VGG-19 [26] and a custom 5 layer CNN (similar to one mentioned in [19]) were used to train the entire USB data set (figure 4) and an external USB 3.0 data set was used to test these trained models. VGG-19 was chosen as the deep learning architecture for this federated application, since it performed the best among all other architectures. Pretrained ImageNet weights were used for the VGG network for training these models. VGG has a total of approximately 139 million parameters out of which 124 millions contribute to its fully connected layers (FC layer). VGG can be used to train a custom data set by fine-tuning the last layer's output equal to the number of labels in the custom data set. For fine-tuned VGG as well as for transfer learning we use the pretrained VGG weights as the initial weights. The pretrained weights also act like a common weight initializer for all the clients taking part in the federated learning, as stated and proved by [19] that models with common initial weights are likely to reach a lower loss gradient and with less CRs.

For the application of disaster classification, [31] applies transfer learning to VGG architecture with a custom classifier layer and trains a model by sharing and federating only the classifier layers' weights. Hence in this research, we tested with both VGG-19 convolution layers as feature extractor, as well as retraining the fine-tuned VGG-19 network for federated learning; the results of these experiments may be seen in the part in section IV. While training the entire VGG-19 model, the total number of parameters to train are around 139 millions, but as the last classifier layer is changed from 1000 to 4 as there are four classes as outputs, the total trainable parameter in our case is approximately 134 millions as shown in figure 7. Whereas the total number of trainable parameters in case of custom fully connected layer for transfer learning are 25,954,564 (aprrox. 26 millions) which can be seen in details in figure 7. It is common knowledge that the greater the number of parameters, the denser the model and the longer it takes to train a model with a high number of parameters. The global federated model acquired by training the entire fine-tuned VGG network (from now on referred to as Process1) is around 550 MB in size, where as the model produced by transfer learning and the weights of only FC layers (further referred as Process2) has model weights size around 140 MB.

### C. Implementation

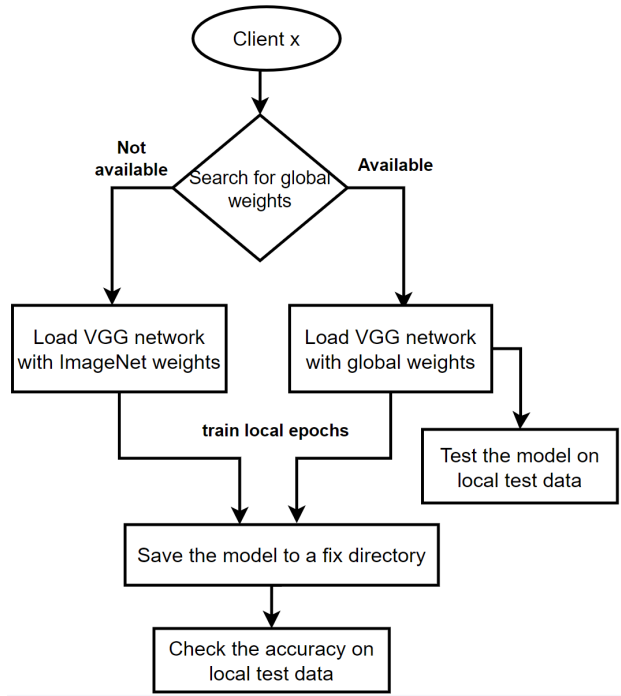

Fig. 8. Client side pipeline

Using TCP-IP protocol within the same network, an application was developed in which three independent computers acted as clients and had access to the server IP where weights were sent and global weights were received. But for faster results and experimentation, we demonstrated that all the three clients and a server were deployed on the same system but had no direct connection and without access to one another's data. This application and experimentation was performed on Ubuntu 20.4 with a 32GB memory Nvidia GeForce RTX 3090 graphics card. To simulate a federated learning scenario, each client's local data was saved independently and was only accessible to that specific client. In our instance, three unique scripts were executed for three distinct clients, and the local weights were stored in a server-accessible location. Figure 8 elaborates more on our custom federated learning workflow on the client side. In accordance with the schematic 8, during the first communication round when there is no global model available in the mentioned path, the client utilises the pretrained ImageNet weights for VGG-19 as the initial weights to train the local epochs. After training, the weights are stored to a designated directory, and then the model's accuracy is evaluated using the test data set. All the clients use the same seed point for common weight initialisation (for fully connected layer initialisation in Process2) and to acquire the same train-test data split on every run. On the server side, once all the clients have finished training the local data, the server creates VGG models equal to the number of clients in the CR to load the client weights and one extra model to store the weights derived from the federated learning algorithm. The federated learning algorithm used for this application is "FedAvg" [19] (as explained in subsection II-B, figure 3) where the resulting global model is achieved by averaging the weights of all the clients and creating a global model which acts as the initial weights for the next communication round for all the clients. After the first communication round when the global weights are available, (as described in figure 8) the client architecture will load these weights and first check the accuracy of this global model on the client's local data and then use them as the initial weights for retraining the local data.

In case of Process1 all VGG weights are stored and federated, but in case of Process2 only the fully connected layers' weights are stored and federated. Successful completion of the resulting global model occurs when the global model achieves the desired result accuracy on all of the client's local test data. By using a custom classifier layer with fewer parameters, as depicted in figure 7, the average time required for a single epoch for all clients in Process1 is approximately 34 seconds, but for Process2 it is just 19.5 seconds.

## IV. Experimental Results and Discussion

The results achieved from global federated model from both Process1 and Process2 are mentioned in table I. Results show that Process1 achieved a far better accurate global model with as less as five communication rounds (CRs). As the communication round increased we noticed that the accuracy of global model on client's local data decreasing and hence we achieved an optimal global model at the 5$^{th}$ CR (the local epochs of each clients was set to 5 epochs). Similarly in case of Process2, as there are less trainable parameters as compared to Process1 (subsection III-B), it required more local epochs as well as more CRs. We experimented with local epochs set to 5, 10, 15 and 20 where as CRs for each of them were taken to around 100 CRs. The best model was always produced around 13$^{th}$ to 16$^{th}$ CR for the respective local epoch number. In this case local epoch set to 10 and 15$^{th}$ CR gave us a good model as compared to other hyper parameters.

TABLE I
ACCURACY OF ALL 3 MODELS ON LOCAL TEST DATA AS WELL AS THE EXTERNAL USB 3.0 DATA SET.

| Accuracy | Normal deep learning with VGG | Federated fine-tuned VGG model (Process1) | Federated transfer learning VGG model (Process2) |
|---|---|---|---|
| Local test data | 99.9% | 99.9% | 99.9% |
| External data | 99.9% | 97.4% | 85% |

**Predicted classification**

| | OKAY | NOT OKAY | HIDDEN | EMPTY |
|---|---|---|---|---|
| OKAY | 2850 | 0 | 1 | 0 |
| NOT_OKAY | 0 | 0 | 0 | 0 |
| HIDDEN | 2 | 0 | 218 | 0 |
| EMPTY | 0 | 0 | 0 | 0 |

Fig. 9. Confusion matrix for classification of external dataset with normal deep learning method using VGG-19 model

**Predicted classification**

| | OKAY | NOT OKAY | HIDDEN | EMPTY |
|---|---|---|---|---|
| OKAY | 2812 | 10 | 15 | 14 |
| NOT_OKAY | 0 | 0 | 0 | 0 |
| HIDDEN | 10 | 25 | 171 | 4 |
| EMPTY | 0 | 0 | 0 | 0 |

Fig. 10. Confusion matrix for classification of external dataset with global federated fine-tuned model (Process1)

| | \multicolumn{4}{c}{Predicted classification} |
|---|---|---|---|---|
| True classification | OKAY | NOT OKAY | HIDDEN | EMPTY |
| OKAY | 2431 | 233 | 167 | 20 |
| NOT_OKAY | 0 | 0 | 0 | 0 |
| HIDDEN | 8 | 17 | 180 | 15 |
| EMPTY | 0 | 0 | 0 | 0 |

Fig. 11. Confusion matrix for classification of external dataset with global federated transfer learning model (Process2)

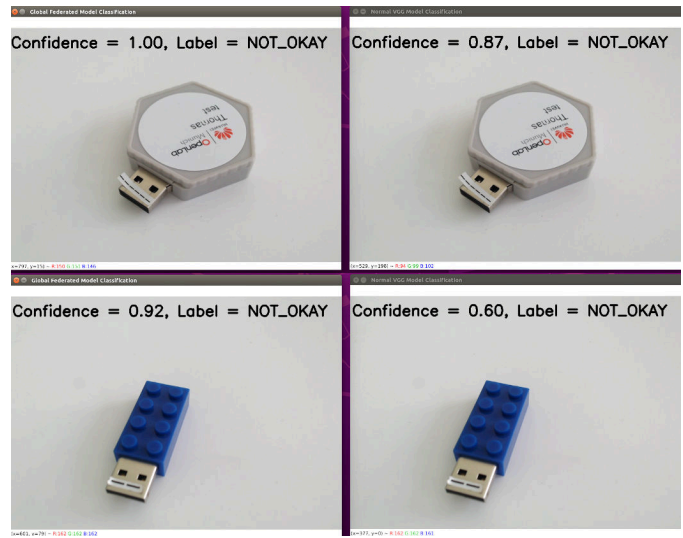

Fig. 12. Live classification of the image using Global federated model(left column) vs. Normal deep learning model (right column)

Figure 9, 10 and 11 illustrates the confusion matrix for classification of the external test data set images (figure 5) by all the three models (3$^{rd}$ model being the normal deep learning model). Similarly, table I demonstrates the best accuracy of all three models on both local and external dataset. For Process2 to achieve the accuracy shown in table I, it required minimum 15 CR. Confusion matrix shown in figure 11 demonstrates that the model cannot perform well on an external test data set. A live classification of USB sticks was done, where the federated transfer learning model (Process2 model) performed the worst. Training a fine-tuned federated VGG model (Process1) outperforms a Transfer learning custom federated model (where both models use pre-trained ImageNet weights) because ImageNet data set [27] does not contain very similar images/objects used for our custom use case. As a result, freezing weights, which acts as a feature extractor in transfer federated learning, did not perform well for quality inspection or manufacturing, as the images are too specific and private to the use case. In contrast, the Process1 model weight size is approximately 500 MB, which can make the transmission of model weights between server and clients challenging. A reference model is trained with the entire data set using normal deep learning methods with the same architecture and with 30 epochs to have a baseline to compare the global federated model. Table I and confusion matrix from figures 9 and 10 illustrate that the model trained with entire data set performs slightly better on the local test set as well as the external data set. But the global federated model (Process1) which is trained by just sharing the model weights between clients, does not lack much behind when compared to the normal deep learning model. To test the robustness of these models, we created a live classification scenario that was slightly different in terms of environment, lighting conditions, and angles than the environment in which the train data set was created.

Figure 12 depicts the classification of a client1 USB stick using fine-tuned federated global model (left side) and normal deep learning (right side). It is interesting to note that client2's USB stick has client1's fault and is appropriately categorised as "NOT_OKAY" by both classification models (lower half). Figure 12 demonstrates that the global federated model is able to classify the image with greater confidence than the model trained using conventional deep learning (with entire data set). This proves that a federated model can perform as well as a standard deep learning model, even without access to raw image data and by just exchanging model weights. In our case the global federated model generalises more effectively than the standard deep learning model as it can be seen by the confidence score from both image classifications as illustrated in figure 12. Due to GDPR law, the primary application of federated learning in Europe is training a global model for a given specific application by several parties/clients, without revealing private or sensitive data.

Figure 13 shows the live classification of this federated use case, in which the global model is compared to each of the three clients trained for 30 epochs on their own local data set, and all of them have the identical architecture (VGG-19 architecture with last layer output as 4). This is to demonstrate that in normal circumstances, each client only has access to it's own data and is limited on training a deep learning model based on its local data. Client3's error USB stick in figure 13 is accurately predicted by client3 and the global federated model. Client1 identifies the USB as "NOT_A_USB" (which is just another term for "EMPTY class") since it has never encountered a USB of that shape. For client2 the error dataset consists of damaged USB sticks and hence it classifies the USB as "OKAY". Figure 14 shows the live classification of client2's USB stick containing client1's error, which was successfully classified by client1's model because the error belongs to its data set, but with a lower confidence score because this USB stick was never in client1's data set. The image is classified as "OKAY" by both client2 and client3 model, since the error does not belong to either clients' data

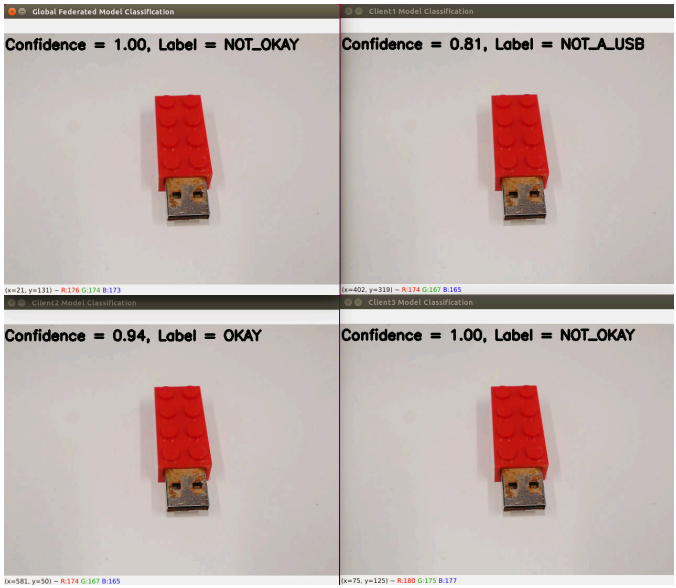

Fig. 13. Live classification of the rusted USB using Global federated model (top left) and all 3 clients trained with respective local data (client 1: top right, client2: bottom left, client3: bottom right); Ground Truth: NOT_OKAY

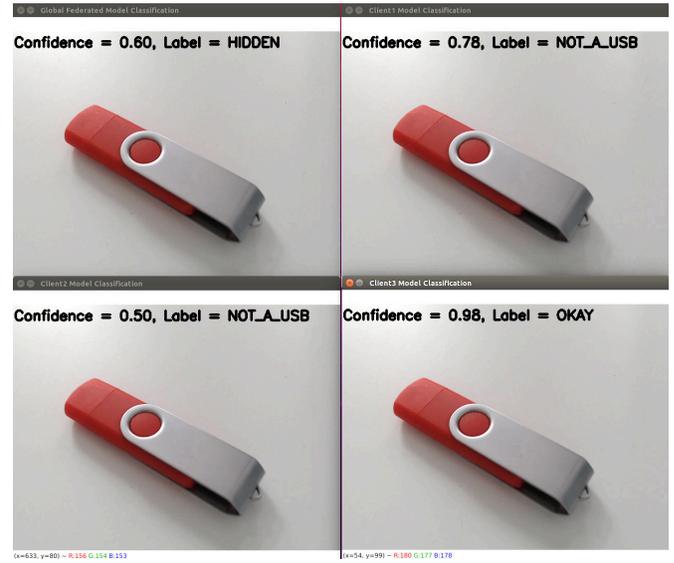

Fig. 15. Live classification of a 3rd party USB using Global federated model (top left) and all 3 clients trained with respective local data (client 1: top right, client2: bottom left, client3: bottom right) Ground Truth: HIDDEN

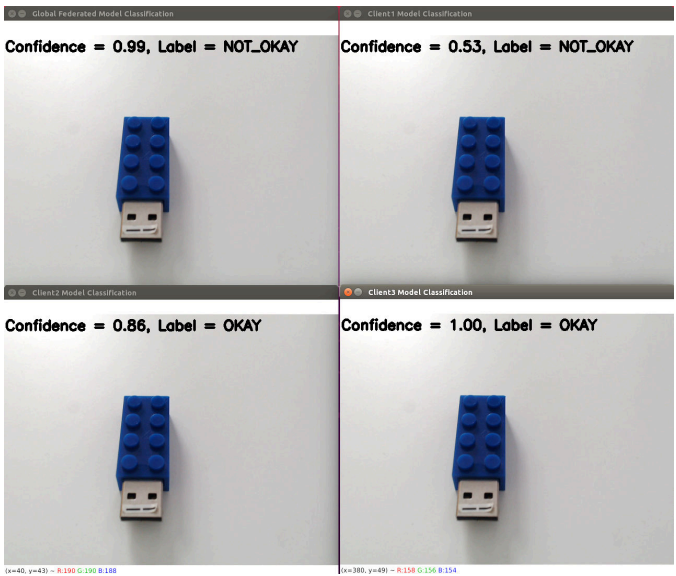

Fig. 14. Live classification of the USB with scratch error (similar to client1's error) using Global federated model (top left) and all 3 clients trained with respective local data (client 1: top right, client2: bottom left, client3: bottom right); Ground Truth: NOT_OKAY

set and remainder of the image seems to be "OKAY" for their respective model, hence also the higher confidence score. From the classification results it is proven that the global model is able to generalise better based on the federation of all the clients' model weights. Hence using federation learning for a quality inspection works well as faults or errors from specific clients can be generalised effectively on other clients' data. Figure 15 shows the testing of a third-party USB stick that was never included in any training data set. It is a unique design which has a movable cap on top that exposes the USB port. This third-party USB stick is likewise accurately classified by the global federated model, which strengthens our use case's reliance on federated learning approach. Federated learning is therefore a useful tool in smart factories if numerous partners wish to construct a quality inspection model for a particular product or use case.

As a part of the experimentation (using Process1), we even tried training the clients with common random initial weights (training from scratch) but it took at least 20 CR to even notice the progress of accuracy of the global model on the clients' local data set and even the global model achieved from $100^{th}$ did not perform well while classifying the external data set (figure 5) or even with the live classification. With random initial weights (different initial weights for all clients) the results where worst (it took 35 CR to even notice a change in local accuracy) than the ones with common initial weights as proved by [19]. Hence using a pretrained ImageNet model for fine-tuned VGG-19 proved to be the best architecture in our scenario.

## V. CONCLUSION

With the implementation of the new GDPR regulations [17], federated learning can be utilised to build a robust deep learning model on private or sensitive data. In this paper we prove that application like quality inspection in manufacturing, where the error/faulty images in production line are few in number and also fewer in variety; using federation learning, multiple parties can collaborate to train a robust quality inspection model for similar products or use case. A method wherein the convolution layers of VGG act as a feature extractor and the weights of custom classifier layers are federated to create a global model (model size 140MB) required at least 15CR to achieve the desired accuracy on an external dataset, but performed poorly in a live classification scenario. Whereas a finely-tuned VGG model (model size 558MB) required only 5 CR to develop a global model that generalised exceptionally well on the USB images. In the live classification scenario, the fine-tuned global federated model detected all errors with a slightly higher confidence score than the VGG model trained on centralised data. Using pre-train ImageNet weights as initial weights for a fine-tune VGG network trained on custom data in a federated context created a classification model with fewer communication rounds. We are amongst the few that have successfully experimented with federated learning using a custom data set in manufacturing, where the global model was able to generalise well despite being constructed by simply exchanging model weights, therefore keeping the clients' local data secure and private.

## VI. Acknowledgement

This project was done in collaboration with Huawei Technologies Duesseldorf GmbH, at the European Research Center in Munich. Thanks for the support of Prof. Juergen Grotepass and Kirill Fridman in this application that also has been demonstrated on Hannover Fair, May 31st 2022.[1]

---

[1] https://www.youtube.com/watch?v=_ZzbiHO846k